# $M^4CD$: A Robust Change Detection Method for Intelligent Visual Surveillance

Kunfeng Wang, *Member, IEEE*, Chao Gou, and Fei-Yue Wang, *Fellow, IEEE*

*Abstract*—In this paper, we propose a robust change detection method for intelligent visual surveillance. This method, named $M^4CD$, includes three major steps. Firstly, a sample-based background model that integrates color and texture cues is built and updated over time. Secondly, multiple heterogeneous features (including brightness variation, chromaticity variation, and texture variation) are extracted by comparing the input frame with the background model, and a multi-source learning strategy is designed to online estimate the probability distributions for both foreground and background. The three features are approximately conditionally independent, making multi-source learning feasible. Pixel-wise foreground posteriors are then estimated with Bayes rule. Finally, the Markov random field (MRF) optimization and heuristic post-processing techniques are used sequentially to improve accuracy. In particular, a two-layer MRF model is constructed to represent pixel-based and superpixel-based contextual constraints compactly. Experimental results on the CDnet dataset indicate that $M^4CD$ is robust under complex environments and ranks among the top methods.

*Index Terms*—Change detection, multimodal background, multi-source learning, conditional independence, Markov random field.

## I. INTRODUCTION

RECENTLY, intelligent visual surveillance has been receiving increased attention in many scientific fields, including computer vision, transportation, healthcare, security and military [1]–[3]. *Change detection* (also referred to as *background subtraction* in some works) is an important early task in these fields. By virtue of change detection, many other applications like object tracking [4], recognition [5], and anomaly identification [6], can be fulfilled.

The basic principle of change detection is to compare the current frame of a video scene with a reference background model, in order to identify zones that are significantly different [7]. Due to environmental complexity in the real world, change detection often encounters a variety of challenges [7]–[10], e.g. dynamic background, camera jitter, intermittent object motion, illumination changes, moving shadows, noise, camouflage, bad weather, low frame rate, camera automatic adjustments, etc.

In practice, it is possible that multiple challenges coexist in a single scene. Unfortunately, very few works are dedicated to addressing the whole set of challenges with a unified framework. Most existing methods suffer from one or more disadvantages:

1) Some methods rely on a single type of feature such as color [11]–[22], edge [23] or texture [24], [25], ignoring the complementary information among features of different types.

2) Some methods [19]–[21], [26]–[28] build a unimodal background model, with the assumption that the background is completely static. However, many dynamic background entities in natural scenes, such as swaying trees and water ripples, violate this assumption.

3) Some methods [11], [13], [15]–[20], [22]–[27], [29]–[32] build adaptive models regarding only the background and recognize foreground pixels purely as outliers. Some other methods [12], [28], [33] assume a uniform distribution for the foreground. As a result, the actual foreground properties in video sequences are discarded, so that camouflaged objects or object-parts can be missed.

4) Some methods [11], [13], [15], [26], [29], [31], [34], [39] improve pixel-wise change detection by using heuristic post-processing techniques such as morphological closing and area filtering. Of course, these techniques can improve performance. On the other hand, as pointed out by [12], when dealing with a probabilistic foreground/background assignment probabilistic methods should be used, such as the use of Markov random fields (MRF). This idea has been verified by many works [18], [21], [32], [37], [38].

In this paper, we propose a robust change detection method which includes three major steps. Firstly, a sample-based multimodal background model that integrates color and texture is maintained over time. Secondly, multiple heterogeneous features (including brightness, chromaticity, and texture variations) are extracted by comparing the current frame with the background model, and a multi-source learning strategy is designed to online estimate the conditional probability distributions for both foreground and background. The term "multi-source" means multiple complementary feature sources about a single physical entity. Multi-source learning exploits the diversity of different sources to discover useful knowledge.

This work was partly supported by National Natural Science Foundation of China (61533019, 71232006).

Kunfeng Wang is with The State Key Laboratory for Management and Control of Complex Systems, Institute of Automation, Chinese Academy of Sciences, Beijing 100190, China, and also with the Innovation Center for Parallel Vision, Qingdao Academy of Intelligent Industries, Qingdao 266000, China (e-mail: kunfeng.wang@ia.ac.cn).

Chao Gou is with The State Key Laboratory for Management and Control of Complex Systems, Institute of Automation, Chinese Academy of Sciences, Beijing 100190, China (e-mail: chao.gou@ia.ac.cn).

Fei-Yue Wang is with The State Key Laboratory for Management and Control of Complex Systems, Institute of Automation, Chinese Academy of Sciences, Beijing 100190, China, and also with the Research Center for Computational Experiments and Parallel Systems Technology, National University of Defense Technology, Changsha 410073, China (e-mail: feiyue@ieee.org).

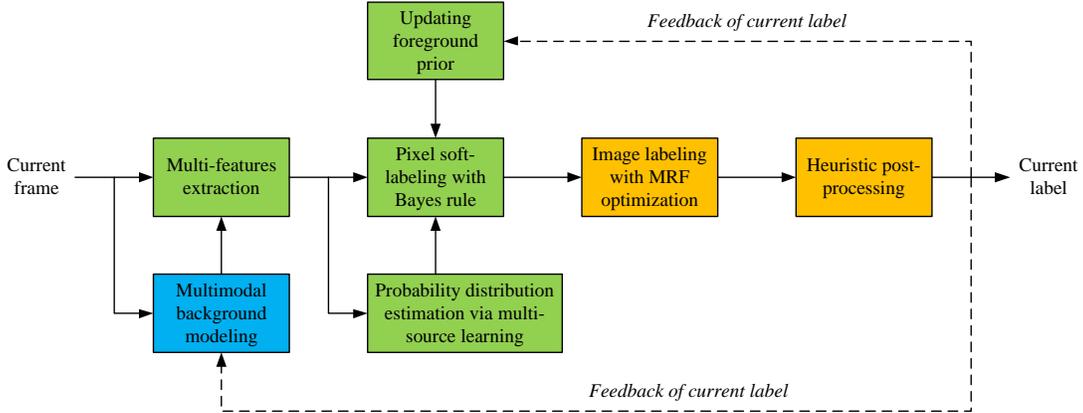

Fig. 1. Flowchart of $M^4CD$. The computation modules of three major steps (corresponding to Sections III–V) are shown in blue, green, and orange respectively.

Pixel-wise foreground posteriors are then estimated with Bayes rule. Finally, optimal image labeling is achieved by combining MRF optimization and heuristic post-processing. In particular, we propose a two-layer MRF model to represent pixel-based and superpixel-based contextual constraints compactly. Considering our method has four important characteristics (i.e., multi-features, multimodal background, multi-source learning, and MRF optimization), we name it as $M^4CD$. To the best of our knowledge, $M^4CD$ is the unique method that integrates all four characteristics. With our method, most challenges in the real world are tackled properly.

This work builds upon our previous work [40]. In that work, we proposed an effective multi-view learning approach to foreground detection for traffic surveillance applications, where multi-features were extracted by comparing the current frame and a reference background image. The multi-source learning strategy in this work is similar to our previous multi-view learning approach [40]. However, in this work we build and update a multimodal background model to facilitate extracting features reliably under dynamic background conditions, which commonly exist in the wild. Moreover, we propose a two-layer MRF model to optimize the foreground/background labeling. Due to these improvements, we are capable of testing the performance of $M^4CD$ on the CDnet dataset [41], which is collected under complex and challenging environmental conditions.

The calculation flowchart of $M^4CD$ is displayed in Fig. 1, where the computation modules of three major steps (corresponding to Sections III–V) are shown in blue, green, and orange respectively. The remainder of this paper is organized as follows. Section II reviews related works. Section III introduces the multimodal background model that integrates color and texture cues. Section IV presents feature extraction and multi-source learning. Section V details the combination of two-layer MRF optimization and heuristic post-processing for image labeling. Experimental results are reported in Section VI, and the conclusion is drawn in Section VII.

## II. RELATED WORKS

The domain of change detection is huge, and a large number of papers have been published in the past decades. So far, there is not a widely approved taxonomy for existent methods. Here, we explore the related works from four aspects.

### A. Features

Features used for change detection include color, edge, texture, motion, etc. The color feature is the most often used [11]–[22], as they are directly available and reasonably discriminative. However, color is susceptible to illumination changes, moving shadows and camouflage. The authors in [24], [25], and [42] used texture features, which are more effective in handling inter-frame illumination changes and shadows. However, the use of texture features can fail to discriminate untextured objects from untextured background. Realizing that a single type of feature has insurmountable limitations, some researchers attempt to integrate multiple complementary features, in order for a stronger robustness in complex environments. For example, Li *et al*. [29] proposed a Bayesian framework that incorporates spectral, spatial, and temporal features to characterize the background appearance. Under that framework, the background is represented by the most significant and frequent features (called *principal features*) at each pixel. Han *et al*. [30] proposed a pixel-wise background modeling and subtraction algorithm using multiple features, where color, gradient, and Harr-like features are integrated to handle spatiotemporal variations for each pixel. Recently, St-Charles *et al*. [31] proposed a change detection method that relies on spatiotemporal binary features as well as color information. This method can ignore most illumination variations and detect camouflaged objects more easily.

### B. Background Model

Basic background models [19]–[21], [26]–[28] assume that feature values of each pixel can be modeled with a unimodal distribution. Those models usually have low complexity, but cannot handle dynamic backgrounds or camera jitter. In fact, multimodal models are more suitable for representing background appearance in real-world scenes. Gaussian mixture models (GMM) [11] and nonparametric kernel density estimation (KDE) [13] are two classical multimodal techniques and still enjoy tremendous popularity today [12], [14], [35], [36]. Nevertheless, Barnich and Van Droogenbroeck [15] argued that there is no imperative to compute the probability density function and invented a universal background subtraction method, called ViBe. This method stores, for each pixel, a set of values taken in the past at the same location or in the neighborhood, and then compares this set to the current pixel value in order to determine whether that pixel belongs to background. ViBe was proved to have high detection rate and low complexity. The sample-based background modeling idea has received considerable concerns [31], [42]. Inspired by ViBe, in this work we maintain a sample-based multimodal background model that integrates color and texture cues.

### C. Foreground Model

Since the appearance of foreground regions is unpredictable, it is difficult to estimate a correct foreground model. Instead, many methods [11], [13], [15]–[20], [22]–[27], [29]–[32], [35], [36] build adaptive models regarding only background and recognize foreground pixels purely as nonmatching points. Some other methods [12], [28], [33] assume a uniform distribution for foreground and use Bayes rule to make the foreground/background decision. Under these simplifications, the actual foreground characteristics in video sequences are not utilized, so that camouflaged objects or background colored object-parts would be missed. A sophisticated foreground model is desired, in order to distinguish between foreground and background in low-contrast images. In [14] and [37], the foreground was characterized by assuming temporal persistence of color and smooth changes in the place of objects. However, in the cases of low frame rate, fast motion and overlapping objects, appropriate temporal information is not available. Benedec and Szirányi [38] proposed to estimate the foreground model by using spatial color information from some pixels in a neighborhood which certainly belong to foreground. Nevertheless, this method leads to a tradeoff between neighborhood size and fidelity: too large a neighborhood reduces the dependence of pixel colors, and too small a neighborhood may lead to very few certain foreground pixels being found in the neighborhood of a pixel. In this paper, we propose a novel foreground modeling idea that relies on the difference between input frames and the background model, rather than on foreground appearance directly.

### D. Regularization

After preliminary foreground/background labels (or probabilities) are assigned to each pixel, regularization is often required to combine information from neighboring pixels and ensure that uniform regions are assigned homogeneous labels. Some methods [11], [13], [15], [26], [29], [31], [34] use heuristic post-processing techniques like closing operation and area filtering. Parks *et al.* [34] evaluated those techniques and found they do improve performance. On the other hand, the authors of [12], [18], [21], [32], [37], and [38] use MRF models to incorporate contextual constraints into the foreground/background decision process and optimize the labeling. However, the existent MRF models for change detection are mainly pixel-based or grid-structured, with each variable/node assigned to one pixel. As a result, the graph structure is very restricted and important relations cannot be modeled. In this paper, we propose a two-layer MRF model to represent pixel-based and superpixel-based constraints compactly.

### E. Our Contributions

Contributions of this paper are summarized as follows:

1) Under the premise that color and texture cues are integrated into a sample-based multimodal background model, multiple heterogeneous features regarding brightness, chromaticity, and texture variations are extracted from the video. These features are robust to dynamic backgrounds and illumination changes.

2) Based on the extracted features, a multi-source learning strategy is designed to online estimate the conditional probability distributions for both foreground and background. These distributions help to better understand foreground and background in the video sequence.

3) After pixel soft-labeling via pixel-wise estimation of foreground posteriors, a two-layer MRF model (composed of a pixel layer and a superpixel layer) is constructed to exploit contextual constrains and optimize the foreground/background segmentation at the frame level.

## III. MULTIMODAL BACKGROUND MODELING

ViBe [15] is a sample-based background subtraction method. In ViBe, some pixel values observed in the past are stored for each pixel. Each pixel value is regarded as a sample, and the union of these samples is regarded as background model. ViBe has many advantages, including *sample-based background representation* and *conservative update*, *memoryless update*, and *spatial propagation* of background samples. In this work, we absorb these advantages. On the other hand, ViBe has some disadvantages. First, it uses only RGB or grayscale values as features, which are less reliable to handle illumination changes and camouflage. Second, it builds only the background model and detects foreground pixels through hard-thresholding. That way, it fails to balance the influences of noise and camouflage. Third, it lacks proper regularization.

In this paper, we extend the ViBe method and try to overcome its disadvantages. The ideas in Section III are inspired by ViBe, while the ideas in Sections IV and V are different from ViBe.

### A. Integration of Color and Texture Cues

In general, color and texture cues are complementary. Color reflects the spectral property at a single pixel, while texture reflects the spatial layout in local region. The integration of them is beneficial. There have been many works [26]–[31], [33], [38] that use both types of cues to detect changes.

In this work, the input frames are in the range of [0, 255]. We maintain a multimodal background model using a sample-based idea inspired by ViBe [15]. A collection of $N$ color samples and $N$ texture samples are stored for each pixel. Each sample corresponds to an observed value taken in the past at the pixel location or in the neighborhood. Formally, the background model $M(x)$ for a pixel $x$ can be expressed as

$$M(x) = \{C_1(x),...,C_i(x),...,C_N(x);T_1(x),...,T_j(x),...,T_N(x)\}, \quad (1)$$

where $C_i(x)$ ( $1 \le i \le N$ ) is a color sample and $T_j(x)$ ($1 \le j \le N$) is a texture sample. The texture samples and color samples are independent, which means that $C_1(x)$ and $T_1(x)$ may be taken from different frames. The number $N$ of samples is a key parameter, which is set to 50 experimentally (see Section VI-B). Based on the sample-based model, any complex background distributions can be represented, without the need to compute probability density functions for each pixel or to estimate statistical parameters, like mean and variance.

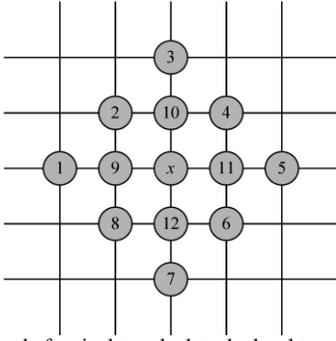

Fig. 2. Neighborhood of a pixel to calculate the local ternary pattern.

In this work, each color sample is represented with an RGB value, but there are many candidates available to represent the texture sample, e.g., local binary pattern [24], local ternary pattern (LTP) [25], [39], local binary similarity pattern [31], and galaxy pattern [42]. Here we use the LTP operator, which has already proved robust to local image noises and illumination changes. In addition, the LTP operator is computationally efficient and compact to describe each pixel using the relative grayscales of its neighboring pixels.

As shown in Fig. 2, if we index the neighborhood of a pixel $x$ by $\{1,...,12\}$, the LTP response can be computed as

$$LTP(x) = \bigoplus_{k=1}^{12} s(I_x, I_k),\qquad(2)$$

where $I_x$ is the grayscale value of the central pixel $x$, $I_k$ is that of its neighboring pixel, $\oplus$ denotes concatenation operator of binary strings, and $s(\cdot)$ is a piecewise function defined as

$$s(I_x, I_k) = \begin{cases} 01, & \text{if } I_k > \max\{(1+\tau)I_x,\ I_x + \upsilon\}, \\ 10, & \text{if } I_k < \min\{(1-\tau)I_x,\ I_x - \upsilon\}, \\ 00, & \text{otherwise}, \end{cases}\qquad(3)$$

where $\tau = 0.1$ is a scale factor indicating the comparison range, and $\upsilon = 5$ is a small tolerance range. According to (3), $(1+\tau)I_x$ takes effect at bright regions (where $I_x \geq 50$), whilst $I_x \pm \upsilon$ takes effect at dark regions (where $I_x < 50$). Here, we distinguish bright regions from dark regions because in dark regions, thermal/dark noise may change the pixel intensity to some degree, so that an additive operator works better. But in bright regions, the noise becomes minor relative to the image intensity, so that a multiplicative operator works better. It should be noted that each comparison results in one of three values, and LTP encodes it with two bits (with "11" undefined). As a result, each texture sample consists of 24 bits.

### B. Model Initialization

In ViBe, the background model was initialized from a single frame. As there is no temporal information available in a single frame, it was assumed that neighboring pixels share a similar temporal distribution. The pixel models were populated with samples selected randomly according to a uniform law in the $3 \times 3$ grid neighborhood of each pixel. A neighboring sample

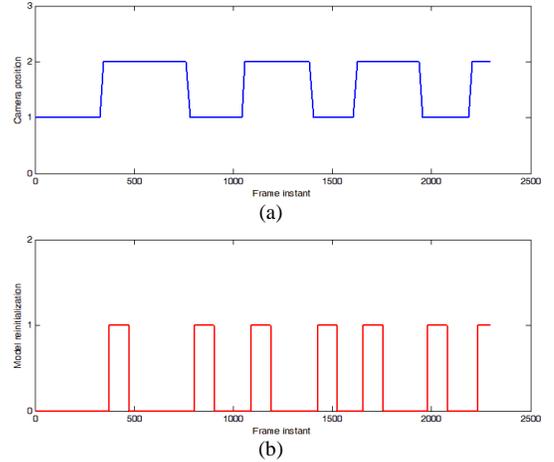

Fig. 3. Example of background model reinitialization for a benchmark video "twoPositionPTZCam". (a) The camera position alternates between "position 1" and "position 2". (b) Frame instants of model reinitialization. It is clear that the model reinitialization is activated timely after the PTZ camera rotates.

can be selected several times or not be selected at all.

In this work, we modify that model initialization strategy slightly. Specifically, we generate a median image with temporal median filtering over the first 100 frames of a video, and then select samples from both the first frame and the median image to initialize the background model. Our new strategy is useful when there are foreground objects in the first frame and helps speed up background recovery in spite of incorrect initialization.

### C. Model Update

As stated earlier, we absorb the advantages of ViBe, including *conservative update*, *memoryless update*, and *spatial propagation* of background samples. Here, we summarize the model update strategy of ViBe and explain our consideration in random subsampling.

In each input frame, a pixel sample can be included in the background model only if it is classified as background. The classification criterion will be presented in Section IV-A. As a result, samples belonging to foreground never have a chance to be included in the background model. This policy guarantees a sharp detection of moving objects, even when the objects are moving slowly.

When updating the background model with a new pixel sample, instead of systematically removing the oldest sample, ViBe selects the sample to be discarded randomly according to a uniform law. The new sample then replaces the selected sample. This policy, called *memoryless update*, offers an exponential decay for the remaining lifespan of the samples.

Without proper processing, a conservative update scheme can lead to deadlock situations and everlasting ghosts. In order to handle these issues, the authors of ViBe proposed a *spatial propagation* policy. Specifically, a new background sample of a pixel is used to update the models of its 8-connected neighboring pixels as well. This leads to a spatial diffusion of information regarding the background evolution. With such a policy, challenges such as dynamic background, camera jitter, and intermittent object motion, can be handled properly.

ViBe uses a *random subsampling* policy. When a pixel value has been classified as background, a random process determines whether that value is used to update the background model. This idea can update the background less frequently and extend the expected lifespan of the background samples. In all its tests, ViBe used a time subsampling factor of 16, i.e., a background pixel value has one chance in 16 of being selected to update the background model. In our implementation, we set the subsampling factor to 1 for the first 100 frames of each video, in order to update the background model more rapidly. Then the subsampling factor is set to 10, until a partial model reinitialization is activated due to the occurrence of drastic background changes, e.g. light switch and PTZ camera rotation. It should be noted that there is no distinction between training frames and test frames. The background model continues being updated whenever a new frame arrives.

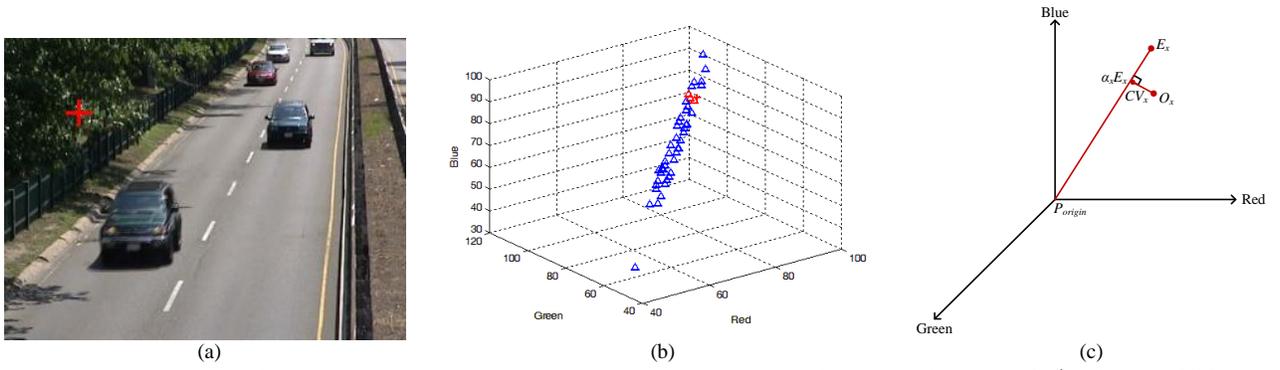

Fig. 4. Illustration of extracting brightness variation and chromaticity variation. (a) An input frame with a pixel marked with red "+". (b) In the RGB color space, the observed pixel value is marked with red "+", the background color samples are marked with "△", and the closest samples are marked with red "△". (c) The extended computational color model.

### D. Model Reinitialization

In practical scenes, it is possible that background changes occur drastically, e.g. when the PTZ camera rotates or indoor lights switch on/off. As a result, we must identify such situations and update the background model properly. We add a frame-level analysis component similar to SuBSENSE [31]. It works by analyzing discrepancies between the background model and the short-term temporal median of input frames.

At each frame instant, the median value of color samples in the background model is computed and then downscaled. The result is denoted as $I_{BG}$. Meanwhile, the temporal median of some recent downscaled input frames is computed and denoted as $I_{TM}$. Then, we compute the disparities between $I_{BG}$ and $I_{TM}$, including the average of color disparities at all pixels ($disp_1$), the percentage of significantly different pixels ($disp_2$), and the exponential entropy measuring the spatial disorder of significantly different pixels ($disp_3$). If the disparities are large enough, i.e., satisfying $disp_1 > 10.0$, $disp_2 > 50\%$ and $disp_3 > 2.65$, we reinitialize the background model partially by adding color and texture samples from the $3 \times 3$ grid neighborhood of each pixel. In addition, we set the subsampling factor to 1 in the following 100 frames, in order to adapt to the background changes rapidly.

Fig. 3 illustrates background model reinitialization for a benchmark video "twoPositionPTZCam" in the CDnet 2014 dataset [41]. Fig. 3(a) shows the camera position alternating between "position 1" and "position 2" aperiodically. Fig. 3(b) shows the frame instants of model reinitialization with the proposed method. As shown, the model reinitialization is activated timely after the PTZ camera rotates.

## IV. FEATURE EXTRACTION AND MULTI-SOURCE LEARNING

In many existent works, only the background model was built and foreground pixels were recognized purely as outliers. That way, it would be difficult to regulate noise and camouflage: too small a threshold will recognize noises falsely as foreground, whilst too large a threshold will recognize camouflaged objects (or object-parts) falsely as background.

In this section, we propose a new idea. Our contributions are to extract multiple heterogeneous features by comparing the current frame with the background model, and then to design a multi-source learning policy to estimate conditional probability distributions regarding both foreground and background.

### A. Feature Extraction

By comparing the current frame with the background model, we extract multiple heterogeneous features, i.e., brightness variation, chromaticity variation, and texture variation. Inspired by [20], we decompose the color variation between the current frame and the background model into brightness variation and chromaticity variation.

#### Extraction of Brightness Variation and Chromaticity Variation

In [20], a computational color model was proposed to extract brightness distortion and chromaticity distortion, but a unimodal background was assumed there. Since multimodal background is common in real-world applications, it is critical to make a multimodal extension for that model.

In the sample-based background model, there is a set of $N$ color samples at each pixel. When estimating color variation between the current pixel observation and the background model, a new value should be compared with only some close samples. Since the background samples often originate from multiple modes, it is reasonable to estimate color variation with a portion of samples rather than with all samples. Let $\#_{close}$ denote the number of close samples. For simplicity, we fix $\#_{close}$ to a small value, i.e., $\#_{close} = 3$.

Given an observed value $O_x$ of any pixel $x$, the $\#_{close}$ closest color samples are picked by sorting the Euclidean distances in the RGB color space. As shown in Fig. 4, for each pixel of the current frame, the distance between its observed value (marked with red "$+$" in Fig. 4(b)) and each background sample is computed, and the $\#_{close}$ closest samples (marked with red "$\triangle$") are picked. In Fig. 4(c), let us denote by $E_x$ one of the closest color samples. For a pair of $O_x$ and $E_x$, brightness variation and chromaticity variation can be computed.

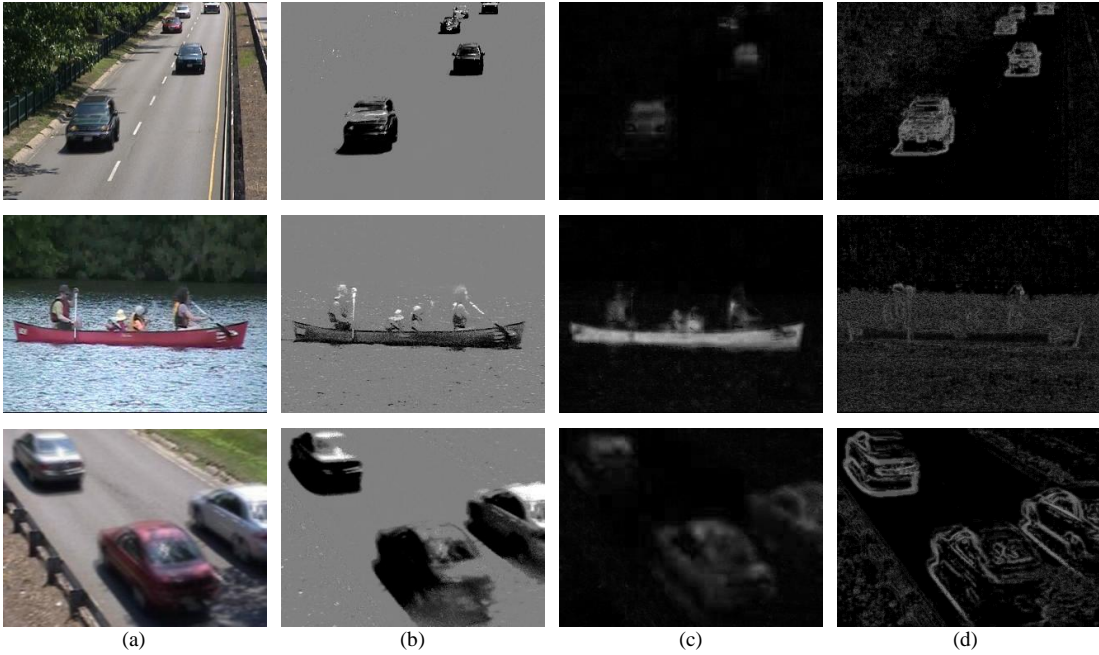

|  (a)  |  (b)  |  (c)  |  (d)  |

Fig. 5. Examples of feature extraction. (a) Input frames. (b) Brightness variations (added by 128). (c) Chromaticity variations (multiplied by 2). (d) Texture variations (normalized to [0,1]). Coming from the CDnet dataset, these videos contain challenges of shadows, dynamic background, and camera jitter.

For any pixel $x$, we first compute $\alpha_x^s$, a ratio between the observed pixel's brightness and the background sample's. Let $O_x = [O_R(x), O_G(x), O_B(x)]$ and $E_x = [E_R(x), E_G(x), E_B(x)]$ denote the RGB color components. According to [20], $\alpha_x^s$ can be computed as

$$\alpha_x^s = \frac{O_R(x)E_R(x) + O_G(x)E_G(x) + O_B(x)E_B(x)}{\left[E_R(x)\right]^2 + \left[E_G(x)\right]^2 + \left[E_B(x)\right]^2}. \qquad (4)$$

In this work, the brightness variation $BV_x^s$ is defined as the signed distance of $\alpha_x^s E_x$ from $E_x$:

$$BV_x^s = \left(\alpha_x^s - 1\right)\left\|P_{origin}E_x\right\|, \qquad (5)$$

where $\left\|P_{origin}E_x\right\|$ denotes the straight-line distance between the origin $P_{origin}$ and the point $E_x$.

As in [20], the chromaticity variation $CV_x^s$ is defined as the orthogonal distance between the observed pixel value $O_x$ and the expected chromaticity line $P_{origin}E_x$:

$$CV_x^s = \sqrt{\left(O_R(x) - \alpha_x^s E_R(x)\right)^2 + \left(O_G(x) - \alpha_x^s E_G(x)\right)^2 + \left(O_B(x) - \alpha_x^s E_B(x)\right)^2}. \qquad (6)$$

Based on $\#_{close}$ pairs of $O_x$ and $E_x$, we obtain multiple $BV_x^s$ and $CV_x^s$. These intermediate results are combined using a median operator to get the final brightness variation $BV_x$ and

chromaticity variation $CV_x$:

$$BV_x = \text{median}_{s \in \#_{close}} BV_x^s,$$
$$CV_x = \text{median}_{s \in \#_{close}} CV_x^s, \qquad (7)$$

where $s \in \#_{close}$ indicates that sample $s$ belongs to the $\#_{close}$ closest color samples.

### Extraction of Texture Variation

Here, we extract texture variation by computing the Hamming distances between the current texture pattern and background texture samples. There has been a collection of $N$ texture samples at each pixel. As done for extracting color variation, we consider only some close texture samples rather than all samples. For any pixel $x$, let us denote by $O_T(x)$ its observed texture value. We can obtain the $\#_{close}$ closest texture samples by sorting the Hamming distances. The Hamming distance between $O_T(x)$ and any close texture sample $E_T(x)$ is denoted by $TV_x^s$. The final texture variation $TV_x$ is obtained via a median operator:

$$TV_x = \text{median}_{s \in \#_{close}} TV_x^s, \qquad (8)$$

where $s \in \#_{close}$ indicates that sample $s$ belongs to the $\#_{close}$ closest texture samples.

We do not use other features such as gradient and optical flow. The gradient feature is used in some works [14], [29], [30], [33]. Gradient belongs to the texture cue, as it reflects the spatial layout at local locations. Using more features means more computations and lower efficiency. Since we have used texture (LTP) variation as feature, we do not use gradient. Optical flow is used by [32], but it cannot be used to handle the CDnet dataset. This is because there is a video category called "*Low Framerate*", in which all videos have low frame rates (less than 1 fps), so that optical flow cannot be computed. This fact prevents us from using optical flow.

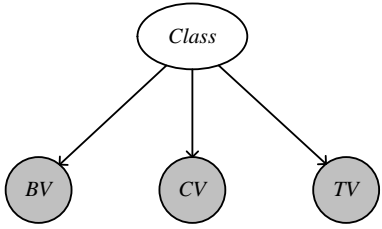

Fig. 6. The naive Bayes model.

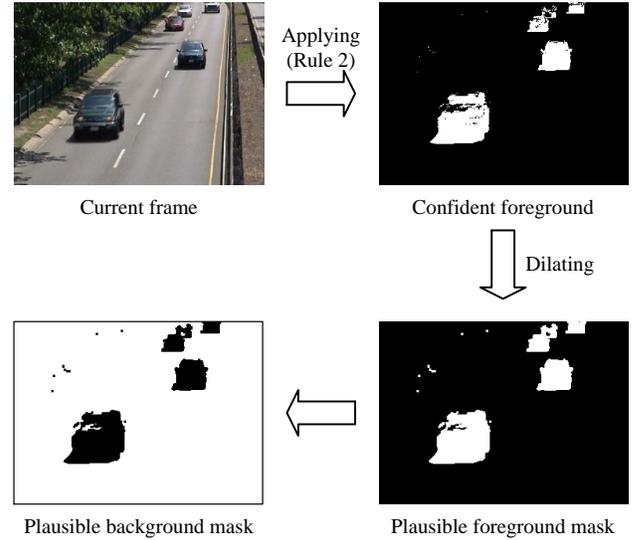

Fig. 7. Diagram to obtain the plausible background mask. In the left-bottom image, the feature values of white pixels are accumulated to estimate the background distribution.

Fig. 5 illustrates the results of feature extraction under some challenges. As shown, the extracted features are reliable under multimodal background conditions. Chromaticity variation and texture variation are insensitive to shadows. In general, background pixels cause low variations in brightness, chromaticity, and texture, whilst foreground pixels cause widespread (but non-uniform) variations in brightness, chromaticity, and texture.

Based on the extracted features, we use Criterion 1 to determine whether to update the background model. The color and texture values of a given pixel $x$ can be used to update background if satisfying

$$\text{abs}(BV_x) < 15 \text{ and } CV_x < 15 \text{ and } TV_x < 8, \quad \text{(Criterion 1)}$$

where $\text{abs}(\cdot)$ denotes the absolute operator. The logic "and" makes this criterion rather strict and conservative. It ensures that new samples must have highly similar appearances to the existent background samples. However, this criterion is not directly used to detect foreground, because pixels that don't satisfy Criterion 1 can also belong to background. In addition to Criterion 1, if a pixel is eventually classified as background (through Section V), it is also regarded as candidate to update the background model.

### B. Conditional Independence of Features

According to their definitions, brightness variation and chromaticity variation are orthogonal in the computational color model. Given the class label *Class* that takes on "FG" (foreground) or "BG" (background), the distribution of chromaticity variation is conditionally independent of brightness variation, and vice versa. In addition, texture variation relies on the spatial layout of a local neighborhood, rather than on the observation at a single pixel. Hence, given the class label, the distribution of texture variation is conditionally independent of brightness variation and chromaticity variation. The conditional independence assumption can be represented using a naive Bayes model of Fig. 6. Based on this model, the joint conditional density can be factorized as

$$p(BV, CV, TV \mid Class) \quad (9)$$
$$= p(BV \mid Class)\, p(CV \mid Class)\, p(TV \mid Class).$$

### C. Probability Distribution Estimation

We have extracted three complementary features, each of which represents a feature source. On top of that, we propose a multi-source learning strategy to estimate conditional probability distributions from real data. In this work, we derive the foreground distribution and the background distribution from global image statistics. Since foreground may never appear in some image regions (e.g., sky), it is impossible to estimate the foreground probability distribution individually for each pixel. Instead, we assume that $BV$, $CV$, and $TV$ have the same statistics in the image. This assumption makes it possible to estimate foreground distributions everywhere.

Due to the conditional independence of features, we have

$$p(BV \mid \text{FG}) = p(BV \mid \text{FG}, CV > \tau_{CV} \text{ or } TV > \tau_{TV}), \quad (10)$$

where $\tau_{CV}$ and $\tau_{TV}$ are thresholds of $CV$ and $TV$, respectively. In other words, given the class label $Class = \text{FG}$, the distribution of brightness variation does not depend on the specific values of chromaticity variation and texture variation.

Furthermore, because background pixels have low chromaticity variation and low texture variation, if $\tau_{CV}$ and $\tau_{TV}$ are large enough, the pixels that satisfy $CV > \tau_{CV}$ or $TV > \tau_{TV}$ can be confidently considered to be foreground pixels. This rule is formulated as

$$\text{If } CV > \tau_{CV} \text{ or } TV > \tau_{TV}, \text{ Then } Class = \text{FG}. \quad \text{(Rule 1)}$$

Combining (10) and (Rule 1), we immediately get

$$p(BV \mid \text{FG}) = p(BV \mid CV > \tau_{CV} \text{ or } TV > \tau_{TV}). \quad (11)$$

The right hand side of (11) leaves out the class label "FG" and indicates that we can use those pixels satisfying $CV > \tau_{CV}$ or $TV > \tau_{TV}$ to estimate $p(BV \mid \text{FG})$.

Similarly, we can estimate the conditional densities of $CV$ and $TV$ given the class label $Class = \text{FG}$,

$$p(CV \mid \text{FG}) = p(CV \mid \text{abs}(BV) > \tau_{BV} \text{ or } TV > \tau_{TV}),$$
$$p(TV \mid \text{FG}) = p(TV \mid \text{abs}(BV) > \tau_{BV} \text{ or } CV > \tau_{CV}), \quad (12)$$

where $\text{abs}(\cdot)$ denotes the absolute operator and $\tau_{BV}$ is a threshold of $BV$. It is critical to set right values for thresholds $\tau_{BV}$, $\tau_{CV}$, and $\tau_{TV}$. They must be large enough, in order to ensure the picked pixels are mainly from foreground. Meanwhile, they should not be too large, otherwise only few pixels are picked out. After careful evaluation on the CDnet dataset, we set $\tau_{BV} = 50$, $\tau_{CV} = 20$, and $\tau_{TV} = 8$.

On the other hand, for estimating the background distribution, we apply (Rule 2) over the current frame to pick "confident" foreground pixels, which are then dilated by using a square structuring element whose width is 3 pixels to propagate the confidence to spatially neighboring pixels and generate a "plausible" foreground mask.

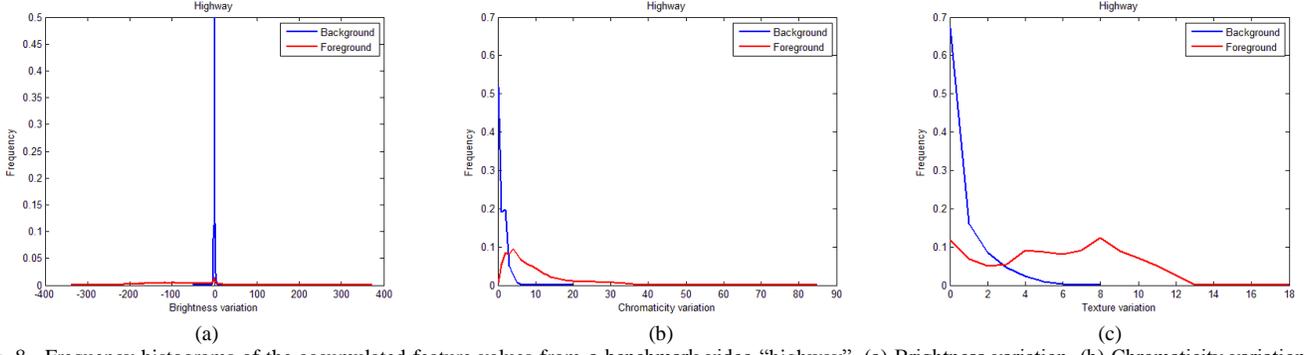

Fig. 8. Frequency histograms of the accumulated feature values from a benchmark video "highway". (a) Brightness variation. (b) Chromaticity variation. (c) Texture Variation. The feature values are picked by multi-source learning. In the charts, blue curves correspond to the background class, and red curves correspond to the foreground class. As shown, the distributions of feature values are rather complex and cannot be approximated well using parametric models.

If $\mathrm{abs}(BV) > \tau_{BV}$ or $CV > \tau_{CV}$ or $TV > \tau_{TV}$, Then $Class = \mathrm{FG}$. (Rule 2)

All the pixels outside the foreground mask constitute a "plausible" background mask, the feature values of which are accumulated to estimate the background distribution. Fig. 7 illustrates the computing diagram, from which it can be seen that most foreground pixels have been excluded from the plausible background mask.

As the system runs, enormous data are accumulated. Fig. 8 shows the histograms of feature values from a benchmark video "highway". These feature values are picked out by our multi-source learning strategy. From Fig. 8, we find that the distributions of feature values are rather complex and cannot be approximated well using parametric models. Hence, we use nonparametric kernel density estimation. Since brightness variation and chromaticity variation are both distances in a certain range, their values can be quantized directly to integers. Texture variation has already an interval of 1.0. As a result, the kernel density estimators require little memory, and the computational cost won't grow with the size of data set. In order to obtain smooth density models, we use Gaussian kernel. The kernel widths for three features are all fixed to 2.0. Because the data set is quite large, the kernel widths can be small.

### D. Pixel Soft-Labeling with Bayes Rule

After probability distribution estimation, we perform pixel soft-labeling with Bayes rule. Specifically, we compute the posteriors of background and foreground given the extracted features. This computation is based on the likelihoods and priors. Given an extracted feature vector $f_x = [BV_x, CV_x, TV_x]$ at any pixel $x$ in the current frame, the posteriors (i.e, soft-labels) of belonging to foreground and background are

$$P_x(\mathrm{FG} \mid f_x) = \frac{p(f_x \mid \mathrm{FG}) \times P_x(\mathrm{FG})}{\sum_{Class \in \{\mathrm{FG}, \mathrm{BG}\}} p(f_x \mid Class) \times P_x(Class)}, \quad (13)$$

$$P_x(\mathrm{BG} \mid f_x) = 1 - P_x(\mathrm{FG} \mid f_x),$$

where $p(f_x \mid Class)$ denotes the likelihood and can be computed based on the naive Bayes model of Fig. 6,

$$p(f_x \mid Class) = p(BV_x \mid Class) p(CV_x \mid Class) p(TV_x \mid Class). \quad (14)$$

It should be noted that the bayes rule of (13) is pixel-based, as we are estimating foreground and background posteriors for each pixel. The likelihood of (14) is based on conditional probability distributions as discussed in Section IV-C, which are estimated based on the entire frame.

$P_x(Class)$ denotes the prior probability of belonging to foreground or background at pixel $x$. In some works [13], [30], the prior was ignored and only the likelihoods were used. However, we believe that an elaborate prior model is essential to higher accuracy. The priors should be spatially distinct. Regions with frequent object motion, such as the road, should have higher foreground priors. The priors should also be time-varying. In recent time, if a pixel position belongs to foreground more frequently than before, its foreground prior should increase; otherwise, its foreground prior should decrease. In light of these, we maintain a dynamic prior model in terms of labels of previous frames,

$$P_{x,t+1}(\mathrm{FG}) = (1 - \rho) P_{x,t}(\mathrm{FG}) + \rho L_{x,t}, \quad (15)$$

where $P_{x,t+1}(\mathrm{FG})$ and $P_{x,t}(\mathrm{FG})$ are pixel $x$'s foreground priors at instants $t+1$ and $t$, respectively. $L_{x,t}$ denotes the pixel $x$'s label at instant $t$, which equals 1 if the pixel is labeled as foreground and equals 0 if labeled as background. $\rho$ is a learning rate, fixed to 0.001 empirically.

In the beginning, the foreground prior $P_{x,0}(\mathrm{FG})$ is initialized to a proper value, such as 0.1 in this work. In the updating stage, $P_{x,t}(\mathrm{FG})$ must not be too low, otherwise occasionally emerging objects will be missed. Formally, we demand

$$P_{x,t}(\mathrm{FG}) = \max \{0.01, P_{x,t}(\mathrm{FG})\}, \quad (16)$$

in order to prevent the foreground prior from becoming too low.

Some examples of pixel soft-labeling are illustrated in Fig. 9. As shown in Fig. 9(b), the foreground priors reflect frequencies of foreground emergences at different locations. From Fig. 9(c), it is clear that true foreground objects have high posteriors to belong to foreground, whereas true background regions have low posteriors to belong to foreground.

## V. Image Labeling with MRF Optimization

Pixel soft-labeling is conducted for each pixel separately,

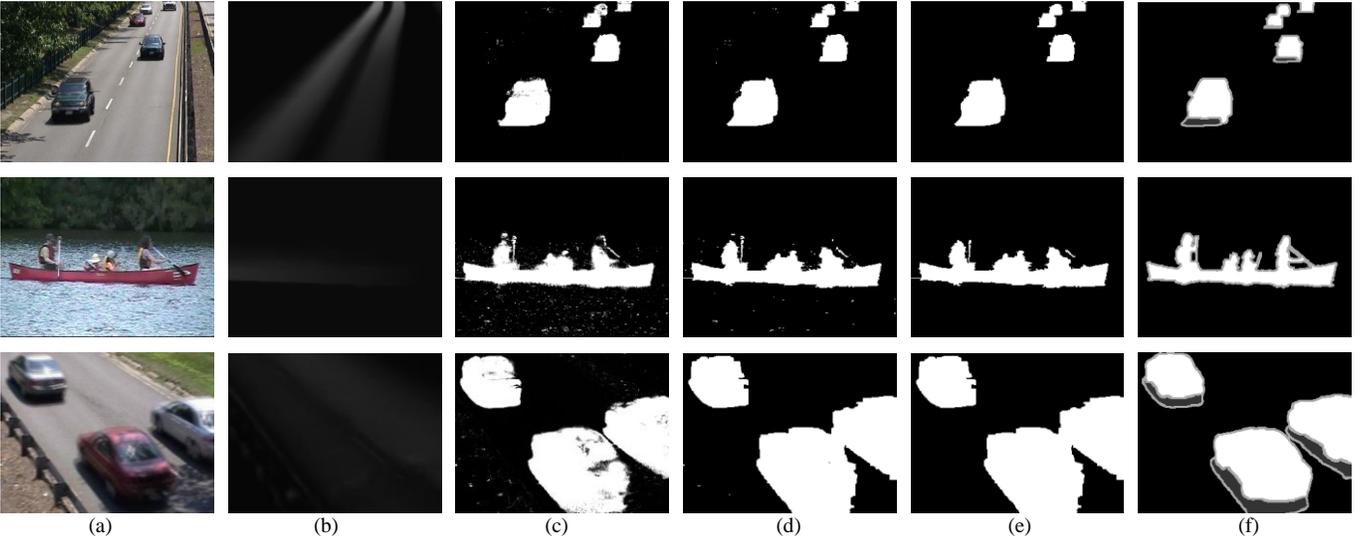

<div style="text-align: center;">(a)        (b)        (c)        (d)        (e)        (f)</div>

Fig. 9. Examples of image labeling. (a) Input frames. (b) Prior probabilities of pixels to belong to foreground. (c) Posterior probabilities of pixels to belong to foreground. The higher the intensity, the more likely the pixel belongs to foreground. (d) Labeling results after MRF optimization. (e) Labeling results after heuristic post-processing. (f) Ground truth. The videos come from the CDnet dataset.

regardless of contextual constraints among neighboring pixels. However, pixel-wise labeling is susceptible to local ambiguity and uncertainty. Hence we use Markov random fields (MRF). In most existing works, a grid-structured MRF is constructed, with a variable/node for each pixel, connected using a four-way spatial neighborhood [12], [18], [21], [32], [37], [38]. Since grid-structured models have limited representation ability, we construct a novel two-layer MRF model to represent pixel-based and superpixel-based constraints compactly.

### A. MRF Optimization

We use the SLIC algorithm [45] to generate superpixels for each input frame, which offers good performance with regard to the tradeoff between boundary adherence and regularity. Fig. 10 shows some images segmented using SLIC.

It seems appealing to use only superpixels to build the MRF model, because the number of nodes and edges can be decreased greatly, and the computational cost is thus reduced. However, two considerations prevent us from doing that. First, the superpixel paradigm usually leads to a coarse segmentation, especially at blurring object boundaries. As a result, accurate object contours are difficult to acquire. Second, under the challenge of intermittent object motion, we expect the sample-based background model to update at the pixel grain and the ghosts to disappear gradually. Superpixel on its own cannot fulfill these requirements, and pixel-based models are also needed. In light of that, we build a two-layer MRF model that integrates pixel-based and superpixel-based constraints. As shown in Fig. 11, our MRF model consists of a pixel layer and a superpixel layer. The explanations of nodes and edges are given in the annotation of Fig. 11.

We are solving a binary labeling problem, where each pixel belongs to foreground or background. Let $X$ be the set of pixels and $SX$ be the set of superpixels in the current frame, and $L$ be the set of labels. The labels are what we want to estimate for each pixel and each superpixel: 1 for foreground and 0 for background. A labeling $l$ assigns a label $l_x \in L$ to each pixel $x \in X$ and a label $l_{sx} \in L$ to each superpixel

$sx \in SX$. Under the MRF framework, the labels should vary slowly almost everywhere but rapidly at some places such as pixels along object boundaries. The quality of a labeling over the whole image is determined by an energy function,

$$
\begin{aligned}
E(l) = &\sum_{x \in X} D(l_x) + \sum_{(x,y) \in NB_4} W(l_x, l_y) \\
&+ \sum_{sx \in SX} C(l_{sx}) + \sum_{(sx,sy) \in NB_{SP}} U(l_{sx}, l_{sy}) \\
&+ \sum_{x \in X, sx \in SX, x \in sx} V(l_x, l_{sx}).
\end{aligned}
\tag{17}
$$

$D(l_x)$ is the data term in the pixel layer, which measures the cost of assigning label $l_x$ to pixel $x$. $NB_4$ is the set of undirected edges in the pixel layer (see Fig. 11(a)). $W(l_x, l_y)$ is the regularization term, which measures the cost of assigning labels $l_x$ and $l_y$ to a pair of neighboring pixels $(x, y)$. $C(l_{sx})$ is the data term in the superpixel layer, which measures the cost of assigning label $l_{sx}$ to superpixel $sx$. $NB_{SP}$ is the set of undirected edges in the superpixel layer (see Fig. 11(b)). $U(l_{sx}, l_{sy})$ is the regularization term, which measures the cost of assigning labels $l_{sx}$ and $l_{sy}$ to a pair of neighboring superpixels $(sx, sy)$. Finally, $V(l_x, l_{sx})$ exerts the compatibility constraint between the label of a superpixel $sx$ and that of its component pixel $x$. A labeling that minimizes the total energy function (17) corresponds to the maximum a posterior estimation of MRF.

In the pixel layer, the data term $D(l_x)$ is given by the pixel-wise posterior probabilities of a pixel belonging to the foreground and to the background:

$$
D(l_x) = \begin{cases} -\log P_x(\text{FG} \mid f_x), & \text{if } l_x = 1, \\ -\log P_x(\text{BG} \mid f_x), & \text{if } l_x = 0, \end{cases}
\tag{18}
$$

where the pixel-wise posteriors have been computed with (13). This term enforces a per-pixel constraint and encourages the labeling to be consistent with per-pixel observation of MRF.

The regularization term $W(l_x, l_y)$ encourages spatial consistency in pixel labels. A cost is paid if two neighboring

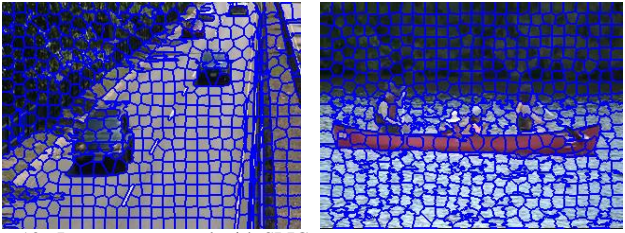

Fig. 10. Images segmented with SLIC.

pixels have different labels. We define the term $W$ as

$$W(l_x, l_y) = \begin{cases} 0, & \text{if } l_x = l_y, \\ \varphi * Z(O_x, O_y), & \text{if } l_x \neq l_y, \end{cases} \quad (19)$$

where $\varphi$ is a weight coefficient. $Z(O_x, O_y)$ is a decreasing function that is controlled by the color difference between pixels $x$ and $y$. In general, the discontinuity of segmentation should coincide with the image discontinuity. We define the function $Z$ as

$$Z(O_x, O_y) = \exp\left(-\frac{\|O_x - O_y\|^2}{\sigma}\right), \quad (20)$$

where $\sigma$ is a parameter which is fixed to 400 empirically, and $\|O_x - O_y\|$ represents the color difference between neighboring pixels $x$ and $y$.

In the superpixel layer, the data term $C(l_{sx})$ is given by the pixel-wise posterior probabilities of all the component pixels:

$$C(l_{sx}) = \begin{cases} \sum_{x \in sx} -\log P_x(\text{FG} \mid f_x), & \text{if } l_{sx} = 1, \\ \sum_{x \in sx} -\log P_x(\text{BG} \mid f_x), & \text{if } l_{sx} = 0, \end{cases} \quad (21)$$

where $x \in sx$ means the pixel $x$ is a component of the superpixel $sx$. This term uses per-pixel observations to pose constraint on the superpixel labeling.

The regularization term $U(l_{sx}, l_{sy})$ encourages spatial consistency in the superpixel labeling. A cost is paid when two neighboring superpixels have different labels. We define the term $U$ as

$$U(l_{sx}, l_{sy}) = \begin{cases} 0, & \text{if } l_{sx} = l_{sy}, \\ \xi, & \text{if } l_{sx} \neq l_{sy}, \end{cases} \quad (22)$$

where $\xi$ is a weight coefficient.

Finally, the compatibility term $V(l_x, l_{sx})$ exists only at the inter-layer edges, i.e., there is a corresponding relation $x \in sx$. It is expected that the labels of two layers are consistent, so a cost is paid when $l_x \neq l_{sx}$. We define the term $V$ as

$$V(l_x, l_{sx}) = \begin{cases} 0, & \text{if } l_x = l_{sx}, \\ \psi, & \text{if } l_x \neq l_{sx}, \end{cases} \quad (23)$$

where $\psi$ is another weight coefficient.

The optimal labeling is found with loopy belief propagation. Although belief propagation is exact only when the graph structure has no loop, in practice it has been proved to be an effective approximate inference technique for general graphical models [43], [44]. In the implementation, we declare convergence when the relative change of messages is less than a threshold $10^{-4}$, thereby obtaining the labeling result of the pixel layer. Some examples of image labeling with MRF optimization are shown in Fig. 9(d).

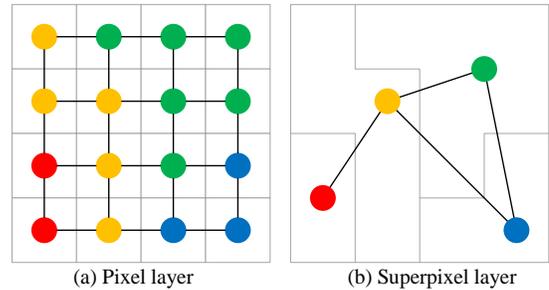

Fig. 11. Two-layer MRF model. (a) In the pixel layer, each node is assigned to one pixel. (b) In the superpixel layer, each node is assigned to a set of pixels forming the superpixel. Solid dots of the same color in (a) and (b) imply the corresponding relation between pixels and the superpixel. While the intra-layer edges are shown in (a) and (b), there are inter-layer edges (not shown here) between each superpixel and its component pixels.

### B. Heuristic Post-processing

Finally, we use heuristic post-processing techniques to improve the change detection result. We first perform area filtering. If a foreground connected region has fewer than $\tau_{Area}$ pixels, it is regarded as detection noise and revised to background. Then, we perform holes filling. If a hole surrounded by foreground pixels has fewer than $\tau_{Area}$ pixels, it is revised to foreground. The value of $\tau_{Area}$ relies on the video resolution. It is set to 25 if the video resolution is less than 2*320*240, and set to 50 otherwise. Some examples of image labeling after heuristic post-processing are shown in Fig. 9(e).

## VI. EXPERIMENTAL RESULTS

### A. Test Dataset and Evaluation Metrics

To conduct experiments on the $M^4CD$, we conduct experiments on the CDnet 2014 dataset [9], which is available at changedetection.net [41]. This dataset consists of 53 video sequences in 11 video categories. As a large-scale dataset composed of real videos, it supplies accurate ground-truth labels and provides a balanced coverage of real-world challenges. In addition, it maintains and updates a rank list of the most accurate change detection algorithms over the years, facilitating algorithm comparison.

The organizers of CDnet proposed to evaluate the ability of a change detection method with seven different metrics: *Recall*, *Specificity*, *False Positive Rate* (FPR), *False Negative Rate* (FNR), *Percentage of Wrong Classifications* (PWC), *F-measure*, and *Precision*. For the *Shadow* category, they also provided an average FPR that is confined to hard-shadow areas (FPR-S) [9], [41]. Here, we use these metrics to evaluate $M^4CD$. We would like to mention that all our change detection results can be downloaded online via the CDnet website, where our method appears under the name "M4CD Version 2.0".

### B. Determination of Parameters

All parameters should be determined experimentally. Due to space limitation, we discuss four of them here: the number $N$ of samples in the background model [see (1)] and the weight coefficients $\varphi$, $\xi$, and $\psi$ in the MRF model [see (19), (22), and (23)]. The other important parameters are discussed in a supplementary file. We use a benchmark video "traffic" from the *Camera Jitter* category to pick these parameters. This video is rather challenging due to the mixture of camera jitter, shadows, and motion blur. The picked parameter values are fixed when processing other videos in the CDnet dataset.

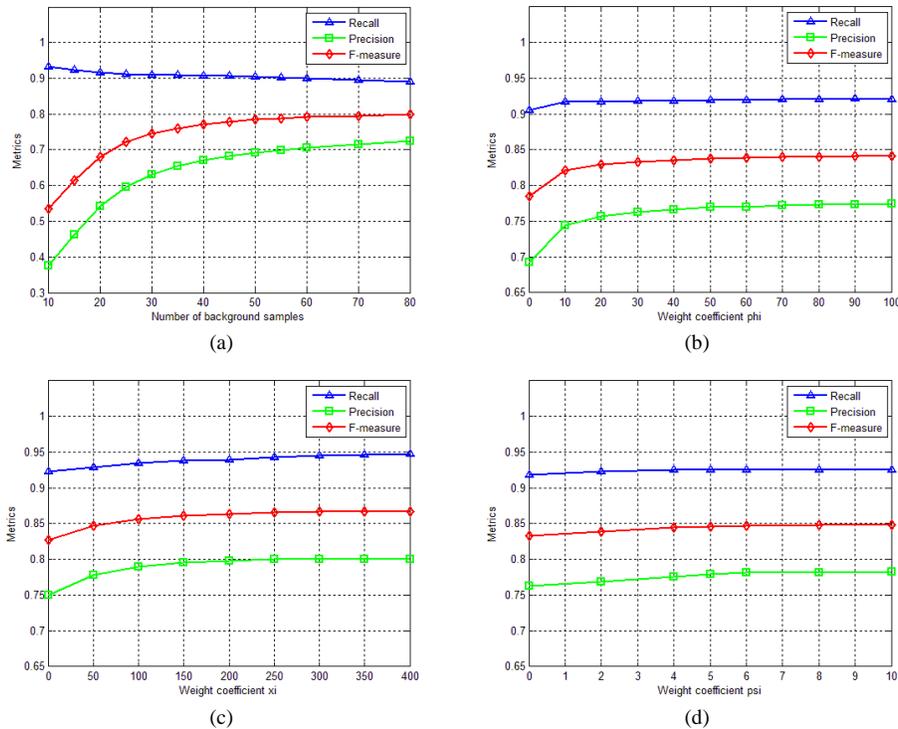

Fig. 12. Curves of recall, precision, and F-measure related to the four critical parameters. The experiments are conducted on a challenging video "traffic".

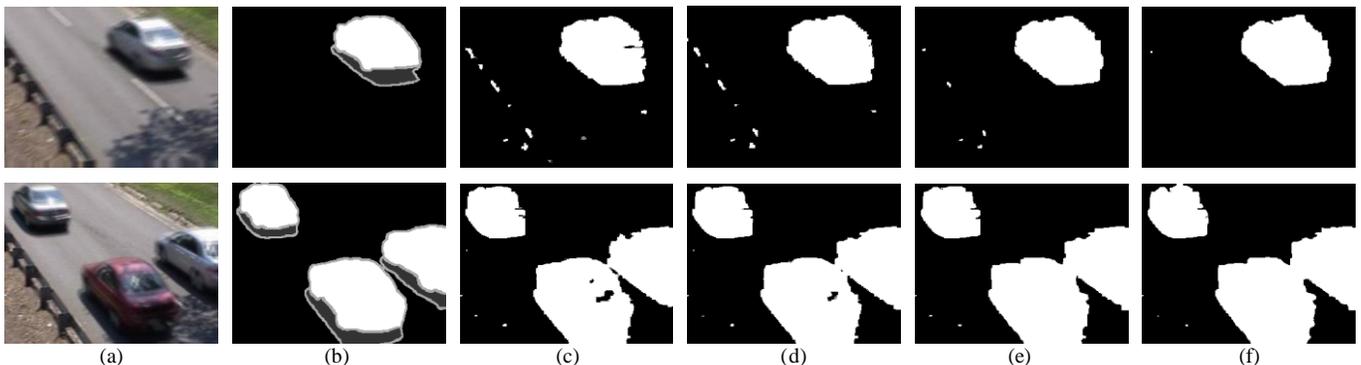

Fig. 13. Illustration of how the compatibility constraint influences the labeling result. (a) Input frames. (b) Ground truth. (c) Labeling results when $\psi$=0. There is no compatibility constraint, which is equivalent to the conventional pixel-layer MRF and leads to many segmentation noises. (d) Labeling results when $\psi$=2. The compatibility constraint is too weak, still leading to many segmentation noises. (e) Labeling results when $\psi$=5. The compatibility constraint is proper. (f) Labeling results when $\psi$=10. The compatibility constraint is too strong, leading to object boundary artifacts.

To select a proper value for $N$ we conduct experiments with $N$ ranging from 10 to 80, estimate per-pixel foreground posteriors and quantify their binary rounding results. Fig. 12(a) shows the curves of recall, precision, and F-measure related to $N$. For this scene, it is impossible to find an $N$ value that optimizes all the metrics simultaneously. The precision and F-measure metrics increase monotonously as $N$ rises, but they tend to saturate for values higher than 50. On the other hand, larger $N$ values induce a lower recall and a greater computational cost. Hence we set $N$=50.

Then, we fix $N$ to 50 and select a proper value for $\varphi$. We consider only the pixel layer in the MRF model, i.e., terms $D$

and $W$ in (17). We conduct experiments with $\varphi$ ranging from 0 to 100, thereby quantifying the pixel-based regularization result. Fig. 12(b) shows the curves of recall, precision, and F-measure related to $\varphi$. As shown, pixel-based regularization can improve the accuracy. But most of the improvements occur when $\varphi$ rises from 0 to 30. Considering that larger $\varphi$ would cause a risk of overfitting, we set $\varphi$ = 30.

Furthermore, we fix $N$ to 50 and select a proper value for $\xi$. Here we consider only the superpixel layer in the MRF model, i.e., terms $C$ and $U$ in (17). We conduct experiments with $\xi$ ranging from 0 to 400, and quantify the superpixel-based regularization result. Fig. 12(c) shows the curves of recall, precision, and F-measure related to $\xi$. Superpixel-based regularization can improve the accuracy. But most improvements occur when $\xi$ rises from 0 to 150. Considering

TABLE I
EVALUATION RESULTS OF $M^4CD$ FOR EACH CATEGORY OF THE CDNET DATASET

| Video category | Recall | Specificity | FPR | FNR | PWC | F-Measure | Precision | FPR-S |
|---|---|---|---|---|---|---|---|---|
| Baseline | 0.9540 | 0.9976 | 0.0024 | 0.0460 | 0.3927 | 0.9322 | 0.9123 | - |
| Dynam. Backg. | 0.8518 | 0.9930 | 0.0070 | 0.1482 | 0.8043 | 0.6857 | 0.6841 | - |
| Camera Jitter | 0.8159 | 0.9921 | 0.0079 | 0.1841 | 1.4478 | 0.8231 | 0.8403 | - |
| Interm. Obj. Motion | 0.7153 | 0.9909 | 0.0091 | 0.2847 | 3.1601 | 0.6939 | 0.8055 | - |
| Shadow | 0.9324 | 0.9922 | 0.0078 | 0.0676 | 1.0796 | 0.8969 | 0.8707 | 0.5749 |
| Thermal | 0.6432 | 0.9981 | 0.0019 | 0.3568 | 2.0839 | 0.7448 | 0.9517 | - |
| Bad Weather | 0.7391 | 0.9990 | 0.0010 | 0.2609 | 0.5037 | 0.8136 | 0.9067 | - |
| Low Framerate | 0.7911 | 0.9949 | 0.0051 | 0.2089 | 0.8394 | 0.6275 | 0.6315 | - |
| Night Videos | 0.6525 | 0.9696 | 0.0304 | 0.3475 | 4.6115 | 0.4946 | 0.4891 | - |
| PTZ | 0.8538 | 0.8984 | 0.1016 | 0.1462 | 10.2247 | 0.2322 | 0.1791 | - |
| Turbulence | 0.7248 | 0.9997 | 0.0003 | 0.2752 | 0.1639 | 0.7978 | 0.8941 | - |
| Overall | 0.7885 | 0.9841 | 0.0159 | 0.2115 | 2.3011 | 0.7038 | 0.7423 | - |

TABLE II
COMPARISON OF $M^4CD$ WITH THE STATE-OF-THE-ART IN TERMS OF OVERALL METRICS

| Method ID | Average ranking across categories | Average ranking | Average Recall | Average Specificity | Average FPR | Average FNR | Average PWC | Average F-Measure | Average Precision |
|---|---|---|---|---|---|---|---|---|---|
| IUTIS-5 [46] | 2.36 | 2.71 | 0.7849 | 0.9948 | 0.0052 | 0.2151 | 1.1986 | 0.7717 | 0.8087 |
| PAWCS [47] | **6.27** | **4.71** | 0.7718 | **0.9949** | **0.0051** | 0.2282 | **1.1992** | 0.7403 | 0.7857 |
| SuBSENSE [31] | 7.55 | 7.57 | **0.8124** | 0.9904 | 0.0096 | **0.1876** | 1.6780 | 0.7408 | 0.7509 |
| SharedModel [48] | 8.55 | 7.00 | 0.8098 | 0.9912 | 0.0088 | 0.1902 | 1.4996 | **0.7474** | 0.7503 |
| FTSG [49] | 8.73 | 9.14 | 0.7657 | 0.9922 | 0.0078 | 0.2343 | 1.3763 | 0.7283 | 0.7696 |
| $M^4CD$ Version 2.0 | 9.27 | 12.29 | 0.7885 | 0.9841 | 0.0159 | 0.2115 | 2.3011 | 0.7038 | 0.7423 |
| SaliencySubsense | 9.36 | 10.43 | 0.7714 | 0.9914 | 0.0086 | 0.2286 | 1.8969 | 0.7176 | 0.7628 |
| * | 9.73 | 10.71 | 0.7416 | 0.9923 | 0.0077 | 0.2584 | 1.8902 | 0.7129 | 0.7754 |
| CwisarDRP | 10.27 | 10.57 | 0.7062 | 0.9947 | 0.0053 | 0.2938 | 1.7197 | 0.7095 | **0.7880** |

* Superpixel Strengthen Background Subtraction.
Note: the best score for each metric is in bold. Because IUTIS-5 is not an independent method, but a combination of 5 top-performing methods, its scores are not in bold.

that larger $\xi$ may cause a risk of overfitting, we set $\xi = 150$.

Finally, we fix other parameters and select a proper value for $\psi$. Here we consider the complete two-layer MRF model. We change $\psi$ from 0 to 10, thereby quantifying the labeling result of the pixel layer. Fig. 12(d) shows the curves of recall, precision, and F-measure related to $\psi$. As shown, the compatibility constraint between the pixel layer and the superpixel layer benefits to higher accuracy. But there is a tradeoff at this place, as illustrated in Fig. 13. The smaller $\psi$, the weaker compatibility constraint, and the more segmentation noises. On the other hand, the larger $\psi$, the stronger compatibility constraint, and the more object boundary artifacts. We find $\psi = 5$ is a proper choice because it results in good balance between noise removal and boundary preservation.

### C. Evaluation Results on CDnet

The entire evaluation results of $M^4CD$ on the CDnet dataset are shown in Table I. It can be seen that $M^4CD$ performs very well for *Baseline* and *Shadow* categories, with F-measures at the level of 0.9. It performs well for *Camera Jitter*, *Bad Weather*, and *Turbulence* categories, with F-measures at the level of 0.8. Nevertheless, *PTZ* and *Night Videos* categories pose heavy challenges to $M^4CD$, with F-measures less than 0.5.

For the *Baseline* category, $M^4CD$ is affected by a video named "PETS2006", where a person keeps motionless in a subway station for a while, causing a number of false negatives. For the *Dynamic Background* category, $M^4CD$ is robust to background motion within a certain range. However, drastic background motion can cause numerous false positives, like in videos "fountain01" and "fall". For the *Camera Jitter* category, $M^4CD$ performs satisfactorily in general, except on the video "boulevard", in which there is a mixture of severe challenges of camera jitter, camouflage, and camera automatic adjustments. Since $M^4CD$ is a bottom-up method and exploits no object-level cues, *Intermittent Object Motion* poses a heavy challenge and cause many false classifications. For the *Shadow* category, $M^4CD$ succeeds in eliminating soft shadows, but is not good at handling hard shadows, with FPR-S as high as 0.5749. For *Thermal* and *Turbulence* categories, even though the video images are in black and white (without chromaticity information available), $M^4CD$ achieves acceptable performance. It is interesting that $M^4CD$ performs well in *Bad Weather* conditions, achieving an F-measure 0.8136. Since no temporal information is relied on by $M^4CD$, *Low Framerate* itself does not become a challenge to this method. However, a video called "port_0_17fps" poses a great challenge due to the mixture of global illumination changes and dynamic background. In *Night Videos*, $M^4CD$ suffers from vehicle headlight reflections on the road and camouflage. *PTZ* camera poses the greatest challenge to $M^4CD$, especially when the camera rotates continuously.

According to statistics at the CDnet website, for the *Turbulence* category, $M^4CD$ ranks second in the state-of-the-art; for *Camera Jitter* and *Thermal* categories, $M^4CD$ ranks third; for the *Shadow* category, $M^4CD$ ranks fourth; and for the *Intermittent Object Motion* category, $M^4CD$ ranks fifth. Table II shows the comparison of $M^4CD$ with the state-of-the-art in terms of overall metrics. The method IDs correspond to methods available at the CDnet website. Note that IUTIS-5 [46] is not an independent method, but a combination of outputs of 5 top-performing methods. Hence, it is reasonable to neglect it during algorithm ranking. Among the state-of-the-art, only PAWCS [47], SuBSENSE [31], SharedModel [48], and FTSG [49] outperform $M^4CD$ in terms of average ranking across categories. But for specific video categories, this is not certainly the case. For instance, $M^4CD$ surpasses PAWCS, SharedModel, and FTSG on the *Turbulence* category, surpasses PAWCS, SuBSENSE, and SharedModel on the *Thermal* category, and surpasses SuBSENSE, SharedModel, and FTSG on the *Camera Jitter* and *Shadow* categories.

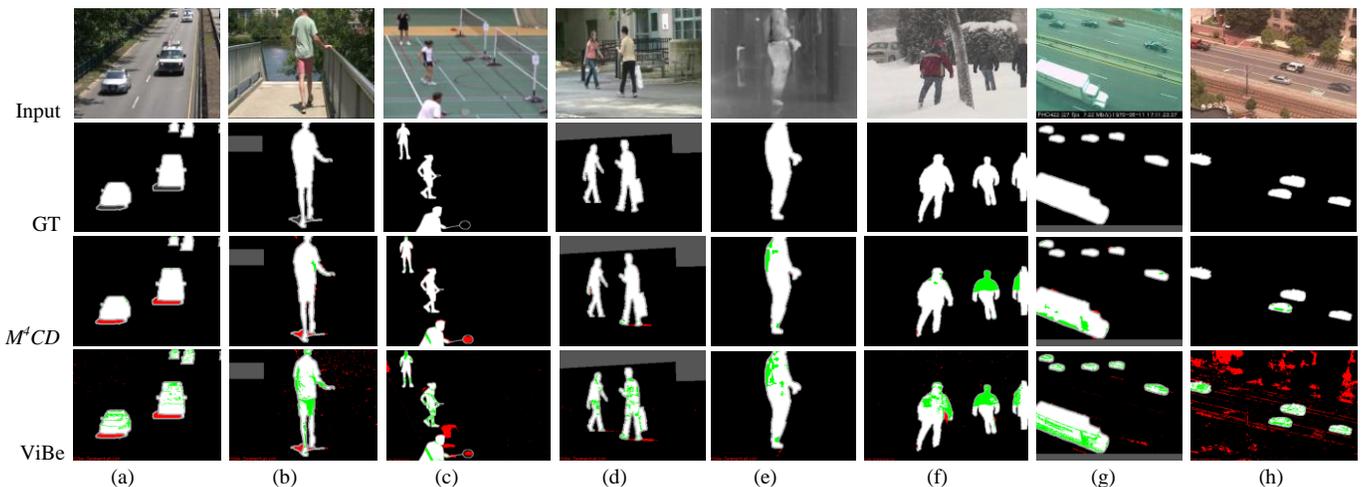

Fig. 14. Typical outputs of $M^4CD$ and ViBe. (a) highway #1514. (b) overpass #2452. (c) badminton #1018. (d) backdoor #1862. (e) corridor #4685. (f) skating #1956. (g) turnpike_0_5fps #1122. (h) intermittentPan #1246. $M^4CD$ results in more accurate foreground/background segmentation than ViBe. Note that we use a color format which adopts the following specifications: true-positive pixels in white, true-negative pixels in black, false-positive pixels in red, and false-negative pixels in green.

TABLE III
METRICS OF EACH PART OF IMAGE LABELING ON "TRAFFIC"

| Part | F-measure | PWC | Precision |
|---|---|---|---|
| With per-pixel likelihoods | 0.6723 | 5.8160 | 0.5178 |
| With per-pixel posteriors | 0.8165 | 2.5301 | 0.7443 |
| With pixel-layer MRF | 0.8409 | 2.1610 | 0.7763 |
| With two-layer MRF | 0.8489 | 2.0500 | 0.7842 |
| After post-processing | **0.8517** | **2.0070** | **0.7888** |

TABLE IV
AVERAGE TIME OF PROCESSING ONE FRAME

| Module | Average processing time (seconds) | |
|---|---|---|
| | 320×240 resolution | 720×480 resolution |
| Multimodal background modeling (Sec. III) | 0.8 | 3.7 |
| Feature extraction and multi-source learning (Sec. IV) | 1.8 | 8.1 |
| Image labeling with MRF optimization (Sec. V) | 2.2 | 9.0 |
| Total | 4.8 | 20.8 |

### D. Evaluating Each Part of $M^4CD$

Figs. 9 and 13 have illustrated some intermediate results of $M^4CD$. Here, we use a benchmark video "traffic" to quantify the contribution of each part of $M^4CD$. Table III displays some metrics corresponding to each part. Comparing the metrics with per-pixel likelihoods and with per-pixel posteriors, there is a large rise in accuracy if the posteriors are used instead of the likelihoods. Based on the per-pixel posteriors, another large rise of accuracy is gained if the two-layer MRF model is used instead of the conventional pixel-layer MRF model. Heuristic post-processing continues improving the accuracy. The three parts of image labeling (Sections IV-D and V) are all important.

### E. Comparison with ViBe

Since $M^4CD$ is inspired by ViBe, we want to compare both methods on the CDnet dataset. Unfortunately, the evaluation results of ViBe are not available at the CDnet website, making it difficult to compare them quantitatively. Here, we use some outputs of $M^4CD$ and ViBe to verify the advantages of $M^4CD$. The inventors of ViBe have publicized their program at their project website [50]. We run that program directly. Fig. 14 shows some typical outputs of $M^4CD$ and ViBe. It is clear that $M^4CD$ results in more accurate foreground/background segmentation than ViBe. Under a variety of environments, the outputs of $M^4CD$ are much closer to the ground truth. Hence, $M^4CD$ outperforms ViBe significantly in terms of accuracy.

### F. Computational Time

$M^4CD$ is implemented on a desktop computer with 2.79GHz Intel Core i7 CPU and 24GB memory. The main program is implemented in MATLAB, with some time-consuming functions (including LTP calculation, feature extraction, and MRF optimization) implemented using C MEX. The computational time is monitored by "tic" and "toc" functions. Table IV shows the average time of processing one frame with two different resolutions 320×240 and 720×480. The cost details of three modules corresponding to Sections III–V are also given. It takes seconds to process one frame, so currently the system cannot work in real-time. In the future, we will use parallel computing platforms to speed up the computation.

## VII. CONCLUSION

In this paper, we propose a robust change detection method called $M^4CD$. This method is inspired by ViBe but has some new characteristics. Both the color and texture cues are integrated into the sample-based background model. Multiple heterogeneous features including brightness, chromaticity, and texture variations are extracted from the video sequence. A multi-source learning strategy is designed to online estimate the conditional probability distributions regarding both foreground and background. This helps to better understand foreground and background in the video sequence. In addition, a two-layer MRF model is presented to optimize the image labeling. Extensive experiments have been conducted on the CDnet dataset, which prove that $M^4CD$ is robust under complex environments and ranks among the top methods. Significantly, it outperforms ViBe in terms of detection accuracy.